\documentclass[conference]{IEEEtran}
\IEEEoverridecommandlockouts
\usepackage{listings}

\usepackage{hyperref}
\usepackage{cite}
\usepackage{amsmath,amssymb,amsfonts}
\usepackage{algorithmic}
\usepackage{graphicx}
\usepackage{textcomp}
\usepackage{array}
\usepackage{xcolor}
\def\BibTeX{{\rm B\kern-.05em{\sc i\kern-.025em b}\kern-.08em
    T\kern-.1667em\lower.7ex\hbox{E}\kern-.125emX}}
\begin{document}

\title{A Deep Learning Framework for Verilog Autocompletion Towards Design and Verification Automation \\
}

\author{\IEEEauthorblockN{Enrique Dehaerne\textsuperscript{1,2}}
\IEEEauthorblockA{\textit{\textsuperscript{1}Faculty of Science, KU Leuven}
 }
\and
\IEEEauthorblockN{Bappaditya Dey\textsuperscript{2} \hspace{28pt} Sandip Halder\textsuperscript{2}}
\IEEEauthorblockA{\textit{\textsuperscript{2}Interuniversity MicroElectronics Centre (IMEC)}
% 3001 Leuven, Belgium
}       
\and
\IEEEauthorblockN{Stefan De Gendt\textsuperscript{2,3}} 
\IEEEauthorblockA{\textit{\textsuperscript{3}Dept. Chemistry, KU Leuven}}
}

\maketitle

\begin{abstract}
Innovative Electronic Design Automation (EDA) solutions are important to meet the design requirements for increasingly complex electronic devices. Verilog, a hardware description language, is widely used for the design and verification of digital circuits and is synthesized using specific EDA tools. However, writing code is a repetitive and time-intensive task. This paper proposes, primarily, a novel deep learning framework for training a Verilog autocompletion model and, secondarily, a Verilog dataset of files and snippets obtained from open-source repositories. The framework involves integrating models pretrained on general programming language data and finetuning them on a dataset curated to be similar to a target downstream task. This is validated by comparing different pretrained models trained on different subsets of the proposed Verilog dataset using multiple evaluation metrics. These experiments demonstrate that the proposed framework achieves better BLEU, ROUGE-L, and chrF scores by 9.5\%, 6.7\%, and 6.9\%, respectively, compared to a model trained from scratch. Code and data are made available at: \url{https://github.com/99EnriqueD/verilog_autocompletion}.
 
\end{abstract}.

\section{Introduction}\label{sec:introduction}

The ever-increasing complexity of electronic systems drives innovation in the electronic design automation (EDA) industry. Verilog is a hardware description language (HDL) used primarily for designing and verifying digital circuits \cite{verilog_wiki} synthesized with EDA tools. Writing custom Verilog modules and functions is often a time-intensive and repetitive task. Rule-based automatic Verilog code generation techniques rely on structured input data and are limited in the types of code they can generate \cite{generating_fpga_cnn_verilog, Automated_ML_Hardware_Synthesis_verilog, verilog_from_cpp,pythonverilog}. Recent advances in machine learning (ML)-based language models, particularly the transformer neural network architecture \cite{attention_is_all}, have led to state-of-the-art (SOTA) results on code generation tasks \cite{codebert, chen2021evaluating, alphacode, codegen}.
% Advances in this field have even enabled commercial products such as the Tabnine\footnote{\url{https://www.tabnine.com/}} and Copilot\footnote{\url{https://github.com/features/copilot/}} autocompletion tools. 
However, these works rely on large datasets of code written in general-purpose programming languages (PLs) that are abundantly available in open-source repository databases \cite{dehaerne2022code_generation_survey}, unlike Verilog.

\begin{figure}
    \centering
    \includegraphics[width=\columnwidth, trim= 0 0 0 0,clip]{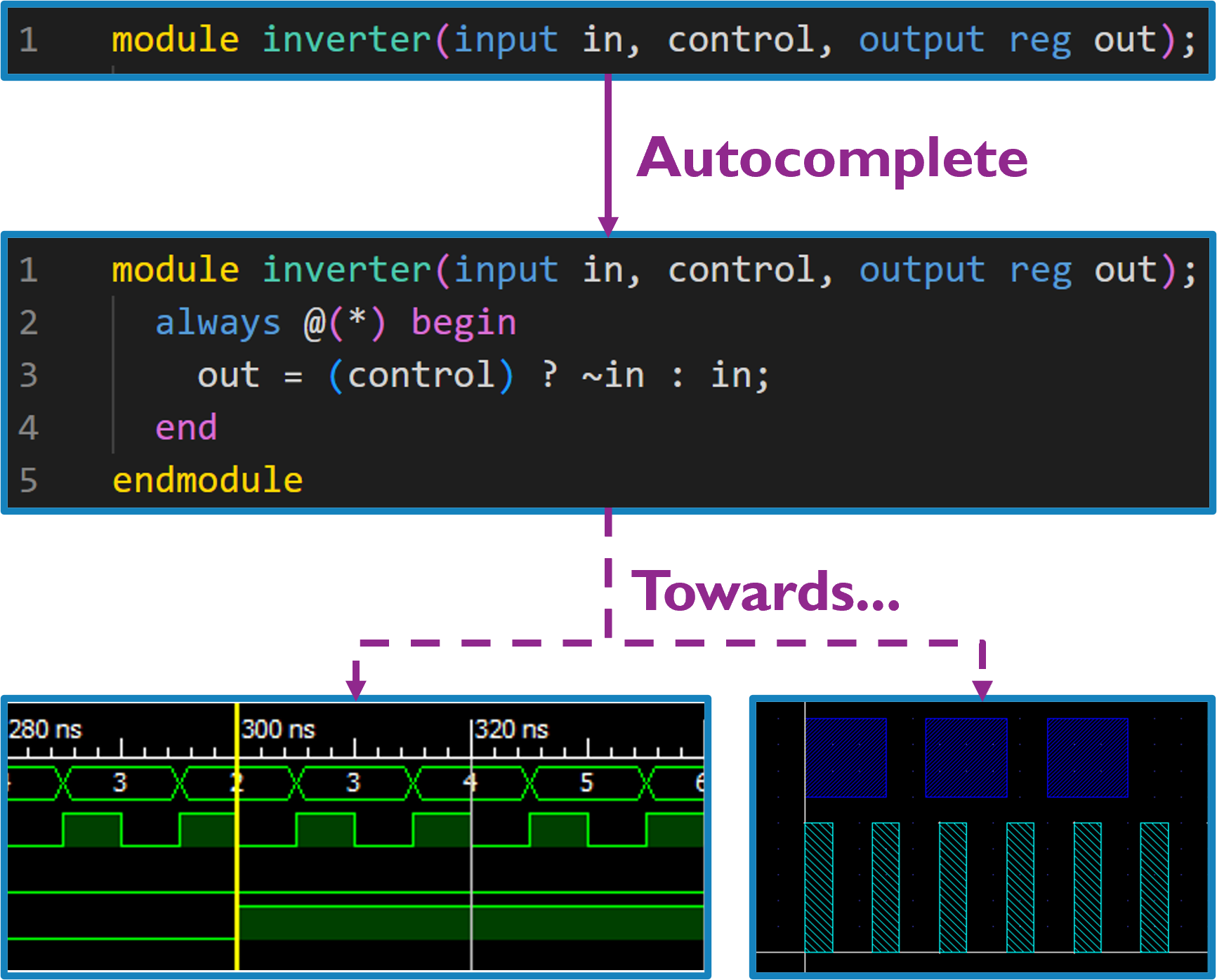}
    \caption{Overview of the problem statement of this study which focuses on autocompleting Verilog modules and functions. This is working towards using the proposed framework to automate design and verification, for example, writing test benches or generating layouts, via automatic Verilog code generation.}
    \label{fig:problem_statement}
\end{figure}

Therefore, this work presents a novel Verilog dataset and a deep learning framework for training models for Verilog code generation tasks. The Verilog dataset was obtained by filtering anomalous files and removing near-duplicate files from open-source repositories. The deep learning framework involves finetuning language models pretrained on large volumes of general PL code on Verilog training data similar to the downstream task and filtered to improve data quality. Experimental results for ML-based autocompletion of Verilog modules and snippets are shown to validate the effectiveness of the proposed framework. This framework is a practical and promising step towards improving the productivity of electronic design engineers via Verilog autocompletion and towards more ambitious downstream EDA tasks such as automating layout generation and test bench synthesis. An example of a Verilog module included in the Verilog dataset is shown in Figure \ref{fig:problem_statement} as well as a graphic depiction of the problem statement described above.

The rest of the paper is organized as follows. The next section details the methodology and results of obtaining a permissive dataset of Verilog files and snippets. Section \ref{sec:deep_learning_framework} presents the proposed deep learning framework to utilize this dataset optimally for training a Verilog code generation model. Section \ref{sec:experimental_design} explains the experiments conducted to verify the proposed framework and Section \ref{sec:results} provides the results of these experiments. Limitations of this study and future work are discussed in Section \ref{sec:limitations_future_work} before concluding the paper.

\section{Related Work}\label{sec:related_work}
The majority of the previous work on automatic Verilog generation is heuristics- or rule-based rather than ML-based. These works essentially translate input data in the form of high-level descriptions \cite{generating_fpga_cnn_verilog, Automated_ML_Hardware_Synthesis_verilog} or other PLs \cite{verilog_from_cpp,pythonverilog} to Verilog code. The advantage of these approaches is that the generated Verilog code is guaranteed to be compilable or synthesizable but is limited in the types of programs they can generate or the types of inputs they can process. Another approach is evolutionary algorithms which can be used to generate novel programs by means of a fitness function without the need for custom heuristics or rules \cite{verilog_evolutionary_generation}. However, this approach requires many fitness function evaluations which can often be very compute-intensive for non-trivial tasks.

ML-based code generation methods can generate code from a wide variety of input data, such as previous-written code or documentation, relatively efficiently at the cost of not guaranteeing compilability or functional correctness. Transformer language models, that process source code as a sequence of tokens, have become especially popular and effective in code generation tasks \cite{dehaerne2022code_generation_survey}. The primary limitation of these methods is the need for a large amount of source code data and substantial computing and specialized hardware to train the models on this data. Most code generation models are trained on huge datasets of code written in general PLs such as Python or Java which are abundantly available on open-source repository databases \cite{dehaerne2022code_generation_survey}. SOTA models are usually trained on a dataset of many different PLs, including the target PL, to maximize the amount of code data the model can learn from \cite{codet5, codegen}. Some studies have shown that the quality of the training data for transformer models is important to get good code generation results \cite{alphacode, google_code_model}. This should be kept in mind when working with open-source code as its quality can be unreliable \cite{dehaerne2022code_generation_survey}.

To the best of our knowledge, only two studies have used ML to generate Verilog code. Pearce et al. \cite{dave_verilog} proposed a model for generating Verilog code from NL descriptions. To do this they used a synthetic dataset of pairs of English descriptions with Verilog code snippets to finetune a GPT-2 model \cite{gpt2}. Contemporaneously to our work, Thakur et al. \cite{thakur2022benchmarking} benchmarked language models pretrained on general PL data for solving Verilog programming challenges. They fine-tuned models on unlabeled Verilog data from books and open-source repositories. The models were functionally evaluated on a collection of Verilog programming problems and corresponding test benches. Our work distinguishes itself from that of Thakur et al. in two ways: (i) we use a larger Verilog dataset comprised only of open-source code repositories to train and evaluate our models and (ii) we propose to curate training data to better finetune for the intended downstream task.

% \section{Proposed Methodology}\label{sec:proposed_methodology}
% In this section, the proposed methodology is introduced and explained. This methodology consists of creating a Verilog Dataset and a deep-learning framework.
\section{Dataset}\label{sec:dataset}
\begin{figure}
    \centering
    \includegraphics[width=\columnwidth, trim= 0 0 0 40,clip]{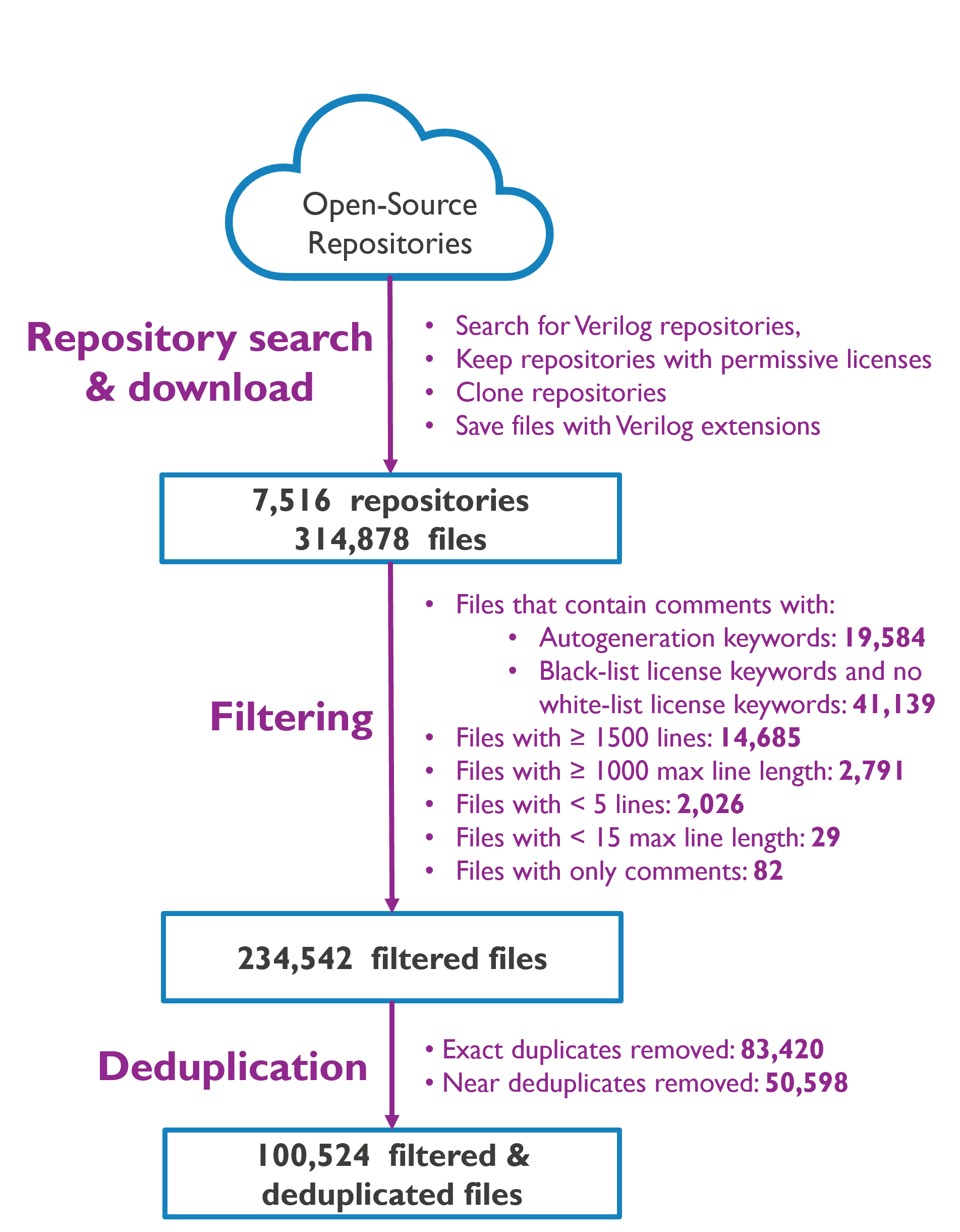}
    \caption{Flowchart of the steps taken to create a reliable Verilog dataset of files with permissive licenses.}
    \label{fig:data_method}
\end{figure}

An integral part of any ML model is the dataset used to train and evaluate it. Due to a lack of publicly available Verilog code datasets, a dataset was created for this study. This dataset consists of two unlabeled subsets, file-level data and snippet-level data, and a labeled subset of snippet definition and body pairs. In this section, the methodology used to create the dataset is explained. An overview of this methodology as well as the number of files at each step is shown in Figure \ref{fig:data_method}. 
% To promote future research into ML for Verilog, this dataset will be made publicly available upon acceptance of this manuscript.

% Dataset creation figure

 An overview of the dataset creation method is shown in Figure \ref{fig:data_method}. First, the GitHub API is used to search for open-source repositories that are identified as containing \textit{Verilog} or \textit{SystemVerilog} files. In the rest of this paper, Verilog and SystemVerilog are treated as different versions of the same language, namely, Verilog. Second, repository licenses were retrieved for all repositories that had them. Then, All repositories with licenses that permit modification and distribution were cloned. Then, Files that did not have Verilog extensions were removed. This resulted in a collection of Verilog files.

% Table of keywords 
% (\textit{v},\textit{verilog},\textit{vlg},\textit{vh},\textit{sv},\textit{svh}, or \textit{svp})

 Inspection of these Verilog files revealed two problematic types of files. The first type of problematic files are autogenerated files. These are problematic since they are highly repetitive and do not reflect human-level Verilog code. The second type of problematic files are files with non-permissive license notices. These files seemingly contradict the license assigned to the repository. Still, an effort was made to remove these files from the dataset to ensure that the dataset respects the permissions given by the developers. Both autogenerated and non-permissive files were removed by means of keyword searches through the comments of the files. All files where at least one autogeneration keyword was found were removed and all files where at least one black-list license keyword and no white-list license keyword were found were removed. 

After applying these previously mentioned filtering steps, still, some files which were likely autogenerated were found. These were mainly large files with many lines and/or very long lines. Through trial-and-error and analysis of the files that were just chosen for filtering, all files with too many lines or a maximum line length that is too long (see Figure \ref{fig:data_method} for threshold values. Very small files were removed as well as these files mostly do not contain useful code. The final filtering step taken was removing files that are empty when comments are removed.

Studies have shown that (near-)duplicates in code datasets have negative effects on code generation models \cite{adverse_effects_duplication, codeparrot_blogpost}. Therefore, exact and near duplicates were removed from the remaining Verilog files. First, exact deduplicates were removed by simply checking the equality of strings read from the Verilog files. For identifying near-deduplicates, the methodology proposed by \cite{adverse_effects_duplication} which calculates the Jaccard similarity between sets of tokens from different files and clusters them if the similarity is above a threshold was used.

The file-level subset of the dataset consists of all remaining files after the previously mentioned processing steps. The snippet-level subset of the dataset consists of all \textit{modules} and \textit{functions} extracted from the file subset using regular expressions. The final, labeled, subset was obtained by splitting snippets into their definition (including their identifier, port list, and return type, where applicable) and body using regular expressions.

Each of the three subsets was divided into three splits: train, validation, and test. To further mitigate the problem of code quality unreliability in open-source repositories \cite{dehaerne2022code_generation_survey} for the evaluation data, only files originating from repositories with at least one \textit{star} on GitHub were eligible for the validation and test splits. Furthermore, since reference outputs are necessary for many automatic evaluation metrics, only files that had at least one snippet extracted from them were eligible for the validation and test splits. Ultimately, 15\% and 35\% of the eligible files were chosen at random for the validation and test splits, respectively. The remaining files deemed eligible for evaluation were assigned to the test set along with the rest of the files. Table \ref{tab:dataset_split} shows the number of files and snippets for each split. The parsable subset of the train split was used for the experimentation of the deep learning framework as explained in the next two sections. As elaborated in Section \ref{sec:experimental_design}, this dataset is small in comparison to the datasets of PLs such as Python which are used by SOTA code generation models like CodeGen \cite{codegen}.

\begin{table}[h]
    \caption{Number of files and snippets in each split of the final dataset.}
    \label{tab:dataset_split}
    \centering
    % \footnotesize
    \begin{tabular}{|c||c|c|c|c|}
        \hline
        \textbf{Split} & \textbf{Files} & \textbf{Snippets} \\
        \hline \hline
        Train (All) &  71,768 & 102,265 \\
        \hline
        Train (Parsable)  & 43,236 & 65,414 \\ 
        \hline
        Validation  & 8,627 & 11,811 \\
        \hline
        Test  & 20129 &  28,207 \\
        \hline \hline
        Total  & 100,524 & 142,283 \\
        \hline
    \end{tabular}
\end{table}

\section{Deep Learning Framework}\label{sec:deep_learning_framework}
\begin{figure}
    \centering
    \includegraphics[width=\columnwidth, trim= 0 0 15 5, clip]{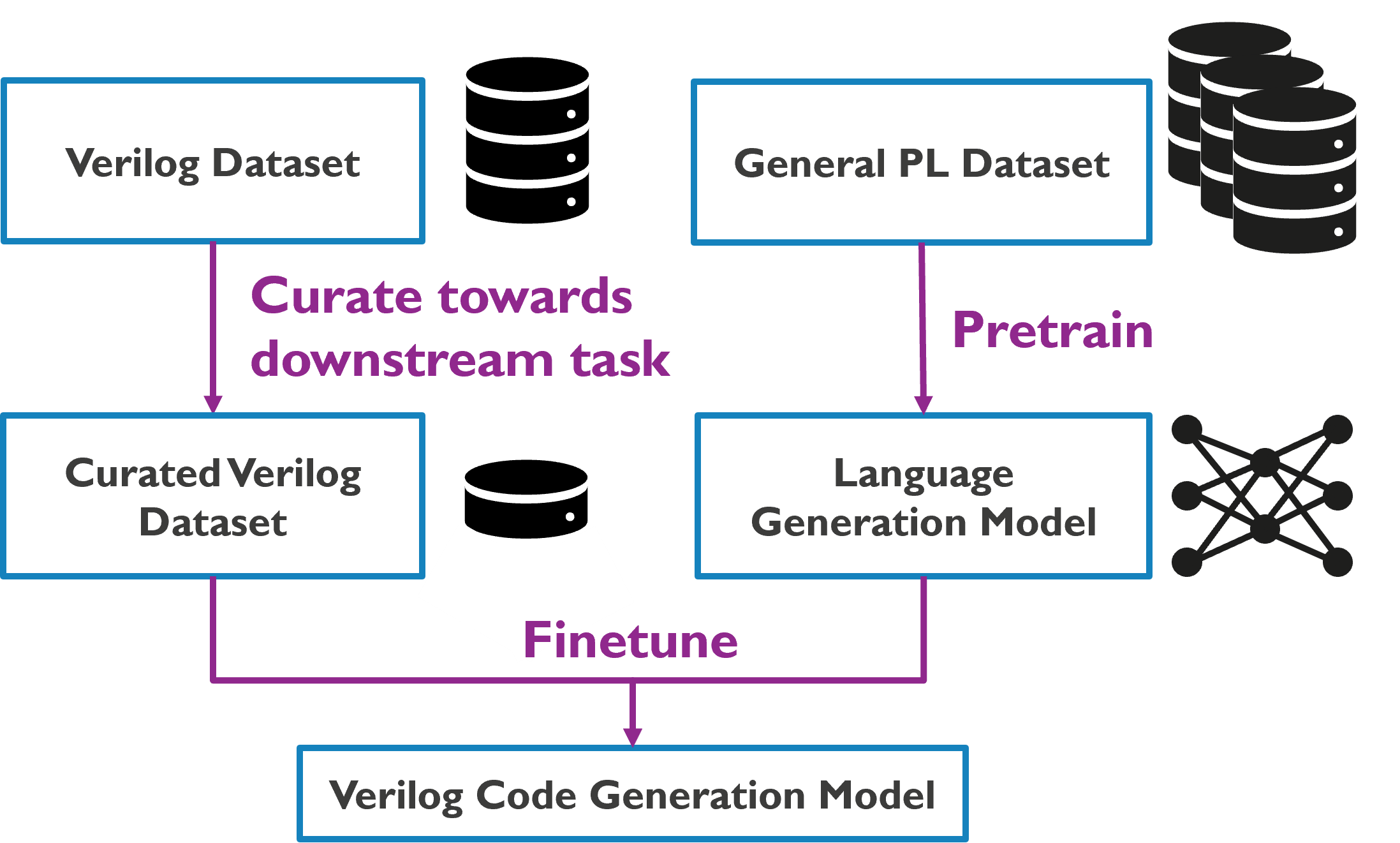}
    \caption{Graphical depiction of the proposed deep learning framework for Verilog autocompletion.}
    \label{fig:dl_framework}
\end{figure}

In this section, the proposed framework to train deep learning models to autocomplete Verilog code is presented. The two main components of the methodology are: (i) using a model pretrained on a large volume of general-purpose programming language data and (ii) finetuning this pretrained model on a dataset of Verilog code curated to be similar to the intended downstream task(s). This framework is graphically depicted in Figure \ref{fig:dl_framework}.

Pretraining of models is especially important when the task-specific dataset is small. Even though Verilog is one of the most popular HDLs, the number of files publicly available is small relative to the number of files of general PLs such as Python or Java. To illustrate this point, the \textit{multi} checkpoint of the SOTA CodeGen model was first trained on \textit{The Pile} \cite{the_pile}, an 800GiB dataset consisting mostly of NL data, and then further trained on 341.1GiB of code from various PLs. The \textit{mono} checkpoint of CodeGen was obtained after further training of the \textit{multi} checkpoint on 217.3GiB of Python-only code. In comparison, all the Verilog dataset code saved in a CSV file is 0.46GiB. Training a model only on a small volume of Verilog code for many epochs will quickly lead to overfitting and degenerated code predictions. Therefore, a model pretrained on data similar to Verilog code is ideal. Until a large-scale HDL dataset is made available, pretraining on general PL data is the best option. 

Once a pretrained model has been chosen, the model must be finetuned to generate Verilog code. Finetuning on general Verilog code is a good option for general-purpose language modeling. However, usually, there is a target downstream task for the model. In this study, autocompleting Verilog modules or functions from their definition is treated as a downstream task different from general Verilog code generation which could include statements usually not found in modules or functions such as package import statements. Inevitably, curating this training dataset for a downstream task will remove data from the general training dataset but it should result in equal or better results on the downstream task using fewer computations during training. Curating can also involve filtering for code quality, for example. 

\section{Experimental Design}\label{sec:experimental_design}
To validate the main components of the proposed framework, experiments were conducted using models pretrained on different datasets and finetuned on different subsets of the Verilog dataset training split. All experiments used the same base architecture, the 350M parameter version of CodeGen \cite{codegen}, a decoder-only transformer neural network \cite{attention_is_all} which has achieved state-of-the-art performance on the HumanEval code generation benchmark \cite{chen2021evaluating}. 

These models were trained using unsupervised autoregressive modeling on 512 token chunks of the given training dataset. They were evaluated in an unlabeled manner on the test split of the file-level data subset using the perplexity metric \cite{perplexity_paper}, a standard metric used for evaluating open-ended generation tasks which were shown to lower (better) as functional evaluation scores improved for different CodeGen models compared \cite{codegen}. To evaluate the generation of modules and functions specifically, the models were evaluated in a labeled manner on the test split of the snippet-level data subset. More concretely, the model being evaluated was given the definition of a snippet and the predicted following tokens generated by the model in a greedy manner are compared to the snippet body as a reference. This comparison is performed using the BLEU \cite{bleu_base}, ROUGE-L \cite{rouge_base}, and chrF \cite{chrf_paper} metrics. The ROUGE-L and chrF metrics were chosen as they were the automatic metrics that aligned best with human judgments for code generation in a recent study \cite{metrics_survey} and the BLEU metric was chosen due to its popularity \cite{dehaerne2022code_generation_survey, metrics_survey}.
To validate that models pretrained on general PLs outperform models initialized from random weights (\textit{scratch}), the \textit{multi} and \textit{mono} checkpoints of the CodeGen models (see Section \ref{sec:deep_learning_framework}) are compared to the same model with random initial weights. To validate that finetuning models using a curated subset of the full dataset outperforms models finetuned on the full dataset, three different subsets of the full training split were used to finetune the \textit{mono} CodeGen model. These subsets were: (i) only snippets, (ii) only files that were parsable by the parsers of either Icarus Verilog \cite{icarus_verilog} or Verilator \cite{verilator}, and (iii) only snippets extracted from files that were parsable. The parsability of a file is used as a proxy metric for code quality. The snippet subset is the code that should be as similar as possible to the supervised evaluation task of autocompleting snippet bodies.

The huggingface library \cite{huggingface} was used for the implementation of the CodeGen models and tokenizer as well as all training and evaluation scripts. All models were trained using four P100 NVIDIA GPUs each with 16GB of memory and a per-device batch size of 4 for one full epoch of the run's respective training dataset. The evaluations were performed using an NVIDIA GeForce RTX 3070 GPU. 

\section{Results \& Discussion}\label{sec:results}
Table \ref{tab:test_results} shows the evaluation results of different models that consist of different combinations of pretrained checkpoints and finetune training data subset. 
For the different pretrained models, the model finetuned from the \textit{mono} checkpoint achieves the best results for all metrics. This shows that the larger the PL dataset of the pretrained model, the better knowledge is transferred to the final model. The most notable improvement is that the perplexity score of the \textit{mono} model achieves a perplexity that is less than half of that achieved by the \textit{scratch} model which suggests that pretraining on PL data is especially useful for general Verilog language modeling. This also results in better snippet autocompletion results as can be shown by the 4.7\%, 4.5\%, and 3.4\% improvements in BLEU, ROUGE-L, and chrF scores, respectively.

\begin{table}[h]
    \caption{Evaluation results on the file-level test split of the Verilog dataset for different models in terms of Perplexity and definition-body pair test split in terms of BLEU, ROUGE-L, and chrF metrics. Lower perplexity scores are better and higher scores of the other metrics are better.}
    \label{tab:test_results}
    \centering
    % Bold the max of each metric
    \begin{tabular}{|>{\centering\arraybackslash}m{33pt}|>{\centering\arraybackslash}m{33pt}||c|c|c|c|}
        \hline
        \textbf{Pretrained Model} & \textbf{Training Data} & \textbf{Perplexity} & \textbf{BLEU} & \textbf{ROUGE-L} & \textbf{chrF} \\ \hline \hline
        scratch & full files & 9166.82 & 0.0953 & 0.1538 & 26.49 \\ \hline
        multi & full files & 4654.38 & 0.0952 & 0.1566 & 27.06 \\ \hline
        mono & full files & \textbf{4387.17} & 0.0998 & 0.1608 & 27.38 \\ \hline \hline
        mono & parsable files & 5625.99 & 0.1031 & 0.1608 & 27.88  \\ \hline
        mono & snippets & 5045.47 & \textbf{0.1044} & \textbf{0.1641} & \textbf{28.31}  \\ \hline
        mono & parsable snippets & 7310.84 & 0.1009 & 0.1615 & 27.85 \\ \hline
    \end{tabular}
\end{table}

Finetuning the \textit{mono} checkpoint only on Verilog snippets, the proposed method, gives the best BLEU, ROUGE-L, and chrF scores of any model. This model achieves improvements on these metrics of 4.6\%, 2.1\%, and 3.4\%, respectively, compared to the same checkpoint finetuned on the full file-level Verilog training subset. This indicates there is a tradeoff between improving downstream task performance using a curated dataset and general Verilog modeling performance using the full Verilog training dataset. Compared to the baseline approach of training a model with random weights on the full training split, the proposed method achieves better BLEU, ROUGE-L, and chrF scores by 9.5\%, 6.7\%, and 6.9\%, respectively. Tables \ref{tab:model_function_example} and \ref{tab:model_module_example} show example outputs of this model on a function and module, sample, respectively, of the snippet-level test split.

\begin{table}[h!]
\caption{Function definition-body pair sample from the test split of the Verilog dataset. The last row shows the prediction from the \textit{mono-snippet} model when given the definition as input. Note: spacing and indents have been modified to improve readability.}
    \label{tab:model_function_example}
    \centering
    % \begin{tabular}{|c||>{\raggedright\arraybackslash}m{300}|}
    \begin{tabular}{|cc|l|}
    \hline
    % \textbf{Input} & \textbf{reference} & \textbf{Prediction}  \\ \hline\hline
     \rotatebox[origin=c]{90}{\textbf{ Definition }} & \rotatebox[origin=c]{90}{\textbf{(Input)}} & 
        \begin{lstlisting}
function void split_string(string str, 
    byte step, ref string result[$]);
        \end{lstlisting}
         \\ \hline
    
    \rotatebox[origin=c]{90}{\textbf{Body}} &
    \rotatebox[origin=c]{90}{\textbf{(Reference)}} &
        \begin{lstlisting}
string tmp_str; 
int i;
bit in_quote;
result = {};
while (i < str.len()) begin
  if (str[i] == "\"") begin
    in_quote = ~in_quote;
  end else if ((str[i] == step) 
        && !in_quote) begin
    result.push_back(tmp_str);
    tmp_str = "";
  end else begin
    tmp_str = {tmp_str, str[i]};
  end
  if (i == str.len()-1) begin
    result.push_back(tmp_str);
  end
  i++;
end
endfunction
        \end{lstlisting} \\ \hline
\rotatebox[origin=c]{90}{\textbf{Prediction}} &
\rotatebox[origin=c]{90}{\textbf{(Mono-Snippet Model)}} & 
        \begin{lstlisting}
bit [31:0] tmp;
int unsigned i;
for (i = 0; i < $size(str); i++) begin
  tmp.push_back({str, i});
  if (!str) begin
    result.push_front(tmp);
  end else if (i == 0) begin
   return result;
  join
end
endfunction
        \end{lstlisting} \\ \hline
    \end{tabular}
\end{table}

\begin{table}[h!]
\caption{Module definition-body pair sample from the test split of the Verilog dataset. The last row shows the prediction from the \textit{mono-snippet} model when given the definition as input. Note: spacing and indents have been modified to improve readability.}
    \label{tab:model_module_example}
    \centering
    % \begin{tabular}{|c||>{\raggedright\arraybackslash}m{300}|}
    \begin{tabular}{|cc|l|}
    \hline
    % \textbf{Input} & \textbf{reference} & \textbf{Prediction}  \\ \hline\hline
     \rotatebox[origin=c]{90}{\textbf{ Definition }} & \rotatebox[origin=c]{90}{\textbf{(Input)}} & 
        \begin{lstlisting}
module clk_divider 
#(parameter WIDTH = 24) (input  clk_in, 
    input  rst_n, output clk_out);
        \end{lstlisting}
         \\ \hline
    \rotatebox[origin=c]{90}{\textbf{Body}} &
    \rotatebox[origin=c]{90}{\textbf{(Reference)}} &
        \begin{lstlisting}
reg [WIDTH-1:0] cnt_div;
always@(posedge clk_in or negedge rst_n)
begin
if(!rst_n)
  cnt_div <= {WIDTH{1'b0}};
else
  cnt_div <= cnt_div + 1'b1;
end
assign clk_out = cnt_div[WIDTH-1];
endmodule
        \end{lstlisting} \\ \hline
\rotatebox[origin=c]{90}{\textbf{Prediction}} &
\rotatebox[origin=c]{90}{\textbf{(Mono-Snippet Model)}} & 
        \begin{lstlisting}
reg [WIDTH-1:0] divisor;
wire [WIDHT-1: 0] divisand;
assign divisand[0] = 1'b0;
assign divider[0]  = ~divisor[0];
always @ (posedge clk_in 
    or negedge rst_n) begin
 if (!rst_ni ||!divisor) begin
  divisor <= {WIDTH{1'b0}};
 end else if (divider[0]) begin
   divisor  <= divisor + 1'b1;
    if (divisor == {WIDHT{1'h0}}) begin
       divisor      <= {WIDH{1'he}};
    end else begin
     divisor   <= {1'b1, divisor};
      if (divider == {WITDH{1'hd}}) begin
        divisor    <= {2'h3, divisand};
          (divide_count > 5'd16) begin
           divisor    <= {divisor, 1'bx};
             else begin
             divisor  <= {~divisor};
        end
     end
end else begin divisand <= {WITHD{1'ha}};
case (divider) 5'd4 : divisornot= 
  {{(WIIH-1){1'h1}}, div};;};[WIDTH+1:1]};
        \end{lstlisting} \\ \hline
    \end{tabular}
\end{table}

Finetuning using the parsable files data subset resulted in relatively smaller improvements in snippet autocompletion metric scores and an even worse perplexity score. This suggests that the potential improvement in data quality by only including parsable code does not compensate for the reduction in the variety of having more data. This is further reflected in the results of the \textit{mono} model finetuned on snippets from parsable files which overall achieves poor perplexity with minimal improvements of the BLEU, ROUGE-L, and chrF scores.

\section{Limitations \& Future Work}\label{sec:limitations_future_work}
The main limitation of this study is the computation-intensive nature of ML-based language models such as CodeGen \cite{codegen}. This means that the smallest model size could only be used for the experiments and training had to be limited to one epoch. The results reported in Table \ref{tab:test_results} could be improved somewhat trivially in future work by using larger CodeGen variants (the largest of which is 16B parameters), as shown by Thakur et al.\cite{thakur2022benchmarking}, and training the model for multiple epochs until the validation loss saturates.

More interesting avenues of future work include using models pretrained on general-purpose PL code as well as other HDL such as VHDL which could be transferred better to Verilog code generation tasks. The relatively low perplexity scores of models pretrained on the full, file-level Verilog training dataset suggests that multi-stage finetuning on increasingly more curated subsets of a dataset could be a data-efficient way of improving downstream task performance. Towards automating layout generation, additional input data such as NL descriptions or structured specifications could be used in combination with code contexts to guide generation. An initial step in this direction could be to associate comments in the Verilog dataset obtained in Section \ref{sec:dataset} with nearby code snippets. 

In the experiments described in Section \ref{sec:experimental_design}, code predictions were generated in a greedy, token-by-token manner. Generation strategies such as top-k sampling \cite{top_k_sampling} could be used to generate many, diverse snippets from the language model. The population of compilable candidate snippets could then be synthesized using EDA tools to provide feedback on the quality of the code snippet. This is related to other works which improved functional correctness metrics by sampling a large number of outputs and used unit tests to filter candidate predictions \cite{chen2021evaluating, alphacode}.

\section{Conclusion}
Increasingly complex electronic circuits have driven the need for innovation in electronic design and verification. Automatic generation of hardware description language code, such as Verilog code, is promising for automating many downstream design and verification tasks such as layout generation. In this paper, a deep-learning framework for training a Verilog code generation model for a target downstream task is proposed. This framework involves using models pretrained on large volumes of general programming language data and finetuning them on Verilog data curated for a target downstream task. To validate this framework, experiments were conducted where this framework is applied to autocompleting Verilog modules and functions. The experiments used a novel, custom Verilog dataset. Promising directions for future work that could improve the proposed framework include augmented pretraining strategies and selecting optimal code predictions.

\section*{Acknowledgment}
We would like to thank our colleagues at IMEC, Victoria Malacara and Dr. Yasser Sherazi, for the discussion and feedback on this project. The resources and services used in this work were provided by the VSC (Flemish Supercomputer Center), funded by the Research Foundation - Flanders (FWO) and the Flemish Government.

\bibliographystyle{IEEEtran}
\bibliography{references/dac_references,references/emnlp_references,references/selected_papers,references/more_survey_references,references/thesis_references}

\begin{thebibliography}{10}
\providecommand{\url}[1]{#1}
\csname url@samestyle\endcsname
\providecommand{\newblock}{\relax}
\providecommand{\bibinfo}[2]{#2}
\providecommand{\BIBentrySTDinterwordspacing}{\spaceskip=0pt\relax}
\providecommand{\BIBentryALTinterwordstretchfactor}{4}
\providecommand{\BIBentryALTinterwordspacing}{\spaceskip=\fontdimen2\font plus
\BIBentryALTinterwordstretchfactor\fontdimen3\font minus
  \fontdimen4\font\relax}
\providecommand{\BIBforeignlanguage}[2]{{%
\expandafter\ifx\csname l@#1\endcsname\relax
\typeout{** WARNING: IEEEtran.bst: No hyphenation pattern has been}%
\typeout{** loaded for the language `#1'. Using the pattern for}%
\typeout{** the default language instead.}%
\else
\language=\csname l@#1\endcsname
\fi
#2}}
\providecommand{\BIBdecl}{\relax}
\BIBdecl

\bibitem{verilog_wiki}
\BIBentryALTinterwordspacing
{Wikipedia contributors}, ``Verilog --- {W}ikipedia{,} the free encyclopedia,''
  2022, accessed: 2022-11-18. [Online]. Available:
  \url{https://en.wikipedia.org/wiki/Verilog}
\BIBentrySTDinterwordspacing

\bibitem{generating_fpga_cnn_verilog}
H.~Zeng, C.~Zhang, and V.~Prasanna, ``Fast generation of high throughput
  customized deep learning accelerators on fpgas,'' in \emph{2017 International
  Conference on ReConFigurable Computing and FPGAs (ReConFig)}, 2017, pp. 1--8.

\bibitem{Automated_ML_Hardware_Synthesis_verilog}
H.~Esmaeilzadeh, S.~Ghodrati, J.~Gu, S.~Guo, A.~B. Kahng, J.~K. Kim, S.~Kinzer,
  R.~Mahapatra, S.~D. Manasi, E.~Mascarenhas, S.~S. Sapatnekar, R.~Varadarajan,
  Z.~Wang, H.~Xu, B.~R. Yatham, and Z.~Zeng, ``Verigood-ml: An open-source flow
  for automated ml hardware synthesis,'' in \emph{2021 IEEE/ACM International
  Conference On Computer Aided Design (ICCAD)}, 2021, pp. 1--7.

\bibitem{verilog_from_cpp}
A.~Madorsky and D.~E. Acosta, ``Vpp - a verilog hdl simulation and generation
  library for c++,'' in \emph{2007 IEEE Nuclear Science Symposium Conference
  Record}, vol.~3, 2007, pp. 1927--1933.

\bibitem{pythonverilog}
S.~Takamaeda-Yamazaki, ``Pyverilog: A python-based hardware design processing
  toolkit for verilog hdl,'' in \emph{Applied Reconfigurable Computing},
  K.~Sano, D.~Soudris, M.~H{\"u}bner, and P.~C. Diniz, Eds.\hskip 1em plus
  0.5em minus 0.4em\relax Cham: Springer International Publishing, 2015, pp.
  451--460.

\bibitem{attention_is_all}
\BIBentryALTinterwordspacing
A.~Vaswani, N.~Shazeer, N.~Parmar, J.~Uszkoreit, L.~Jones, A.~N. Gomez,
  L.~Kaiser, and I.~Polosukhin, ``Attention is all you need,'' \emph{CoRR},
  vol. abs/1706.03762, 2017. [Online]. Available:
  \url{http://arxiv.org/abs/1706.03762}
\BIBentrySTDinterwordspacing

\bibitem{codebert}
\BIBentryALTinterwordspacing
Z.~Feng, D.~Guo, D.~Tang, N.~Duan, X.~Feng, M.~Gong, L.~Shou, B.~Qin, T.~Liu,
  D.~Jiang, and M.~Zhou, ``Codebert: {A} pre-trained model for programming and
  natural languages,'' \emph{CoRR}, vol. abs/2002.08155, 2020. [Online].
  Available: \url{https://arxiv.org/abs/2002.08155}
\BIBentrySTDinterwordspacing

\bibitem{chen2021evaluating}
M.~Chen, J.~Tworek, H.~Jun, Q.~Yuan, H.~P. de~Oliveira~Pinto, J.~Kaplan,
  H.~Edwards, Y.~Burda, N.~Joseph, G.~Brockman, A.~Ray, R.~Puri, G.~Krueger,
  M.~Petrov, H.~Khlaaf, G.~Sastry, P.~Mishkin, B.~Chan, S.~Gray, N.~Ryder,
  M.~Pavlov, A.~Power, L.~Kaiser, M.~Bavarian, C.~Winter, P.~Tillet, F.~P.
  Such, D.~Cummings, M.~Plappert, F.~Chantzis, E.~Barnes, A.~Herbert-Voss,
  W.~H. Guss, A.~Nichol, A.~Paino, N.~Tezak, J.~Tang, I.~Babuschkin, S.~Balaji,
  S.~Jain, W.~Saunders, C.~Hesse, A.~N. Carr, J.~Leike, J.~Achiam, V.~Misra,
  E.~Morikawa, A.~Radford, M.~Knight, M.~Brundage, M.~Murati, K.~Mayer,
  P.~Welinder, B.~McGrew, D.~Amodei, S.~McCandlish, I.~Sutskever, and
  W.~Zaremba, ``Evaluating large language models trained on code,'' 2021.

\bibitem{alphacode}
\BIBentryALTinterwordspacing
Y.~Li, D.~Choi, J.~Chung, N.~Kushman, J.~Schrittwieser, R.~Leblond, T.~Eccles,
  J.~Keeling, F.~Gimeno, A.~D. Lago, T.~Hubert, P.~Choy, C.~d.~M. d'Autume,
  I.~Babuschkin, X.~Chen, P.-S. Huang, J.~Welbl, S.~Gowal, A.~Cherepanov,
  J.~Molloy, D.~J. Mankowitz, E.~S. Robson, P.~Kohli, N.~de~Freitas,
  K.~Kavukcuoglu, and O.~Vinyals, ``Competition-level code generation with
  alphacode,'' 2022. [Online]. Available:
  \url{https://arxiv.org/abs/2203.07814}
\BIBentrySTDinterwordspacing

\bibitem{codegen}
E.~Nijkamp, B.~Pang, H.~Hayashi, L.~Tu, H.~Wang, Y.~Zhou, S.~Savarese, and
  C.~Xiong, ``Codegen: An open large language model for code with multi-turn
  program synthesis,'' \emph{arXiv preprint}, 2022.

\bibitem{dehaerne2022code_generation_survey}
E.~Dehaerne, B.~Dey, S.~Halder, S.~De~Gendt, and W.~Meert, ``Code generation
  using machine learning: A systematic review,'' \emph{IEEE Access}, vol.~10,
  pp. 82\,434--82\,455, 2022.

\bibitem{verilog_evolutionary_generation}
\BIBentryALTinterwordspacing
A.~A. Vivekananda and E.~Enoiu, ``Automated test case generation for digital
  system designs: A mapping study on vhdl, verilog, and systemverilog
  description languages,'' \emph{Designs}, vol.~4, no.~3, 2020. [Online].
  Available: \url{https://www.mdpi.com/2411-9660/4/3/31}
\BIBentrySTDinterwordspacing

\bibitem{codet5}
\BIBentryALTinterwordspacing
Y.~Wang, W.~Wang, S.~Joty, and S.~C.~H. Hoi, ``Codet5: Identifier-aware unified
  pre-trained encoder-decoder models for code understanding and generation,''
  2021. [Online]. Available: \url{https://arxiv.org/abs/2109.00859}
\BIBentrySTDinterwordspacing

\bibitem{google_code_model}
\BIBentryALTinterwordspacing
M.~Tabachnyk and S.~Nikolov, ``Ml-enhanced code completion improves developer
  productivity,'' July 2022, accessed: 2022-11-19. [Online]. Available:
  \url{https://ai.googleblog.com/2022/07/ml-enhanced-code-completion-improves.html}
\BIBentrySTDinterwordspacing

\bibitem{dave_verilog}
H.~Pearce, B.~Tan, and R.~Karri, ``Dave: Deriving automatically verilog from
  english,'' in \emph{2020 ACM/IEEE 2nd Workshop on Machine Learning for CAD
  (MLCAD)}, 2020, pp. 27--32.

\bibitem{gpt2}
A.~Radford, J.~Wu, R.~Child, D.~Luan, D.~Amodei, and I.~Sutskever, ``Language
  models are unsupervised multitask learners,'' 2019.

\bibitem{thakur2022benchmarking}
S.~Thakur, B.~Ahmad, Z.~Fan, H.~Pearce, B.~Tan, R.~Karri, B.~Dolan-Gavitt, and
  S.~Garg, ``Benchmarking large language models for automated verilog rtl code
  generation,'' 2022.

\bibitem{adverse_effects_duplication}
\BIBentryALTinterwordspacing
M.~Allamanis, ``The adverse effects of code duplication in machine learning
  models of code,'' in \emph{Proceedings of the 2019 ACM SIGPLAN International
  Symposium on New Ideas, New Paradigms, and Reflections on Programming and
  Software}, ser. Onward! 2019.\hskip 1em plus 0.5em minus 0.4em\relax New
  York, NY, USA: Association for Computing Machinery, 2019, p. 143–153.
  [Online]. Available: \url{https://doi.org/10.1145/3359591.3359735}
\BIBentrySTDinterwordspacing

\bibitem{codeparrot_blogpost}
\BIBentryALTinterwordspacing
L.~Von~Werra, ``Training codeparrot from scratch,'' dec 2021. [Online].
  Available: \url{https://huggingface.co/blog/codeparrot}
\BIBentrySTDinterwordspacing

\bibitem{the_pile}
\BIBentryALTinterwordspacing
L.~Gao, S.~Biderman, S.~Black, L.~Golding, T.~Hoppe, C.~Foster, J.~Phang,
  H.~He, A.~Thite, N.~Nabeshima, S.~Presser, and C.~Leahy, ``The pile: An 800gb
  dataset of diverse text for language modeling,'' 2021. [Online]. Available:
  \url{https://arxiv.org/abs/2101.00027}
\BIBentrySTDinterwordspacing

\bibitem{perplexity_paper}
F.~Jelinek, R.~L. Mercer, L.~R. Bahl, and J.~K. Baker, ``Perplexity—a measure
  of the difficulty of speech recognition tasks,'' \emph{The Journal of the
  Acoustical Society of America}, vol.~62, no.~S1, pp. S63--S63, 1977.

\bibitem{bleu_base}
\BIBentryALTinterwordspacing
K.~Papineni, S.~Roukos, T.~Ward, and W.-J. Zhu, ``{B}leu: a method for
  automatic evaluation of machine translation,'' in \emph{Proceedings of the
  40th Annual Meeting of the Association for Computational Linguistics}.\hskip
  1em plus 0.5em minus 0.4em\relax Philadelphia, Pennsylvania, USA: Association
  for Computational Linguistics, Jul. 2002, pp. 311--318. [Online]. Available:
  \url{https://aclanthology.org/P02-1040}
\BIBentrySTDinterwordspacing

\bibitem{rouge_base}
\BIBentryALTinterwordspacing
C.-Y. Lin, ``{ROUGE}: A package for automatic evaluation of summaries,'' in
  \emph{Text Summarization Branches Out}.\hskip 1em plus 0.5em minus
  0.4em\relax Barcelona, Spain: Association for Computational Linguistics, Jul.
  2004, pp. 74--81. [Online]. Available:
  \url{https://aclanthology.org/W04-1013}
\BIBentrySTDinterwordspacing

\bibitem{chrf_paper}
\BIBentryALTinterwordspacing
M.~Popovi{\'c}, ``chr{F}: character n-gram {F}-score for automatic {MT}
  evaluation,'' in \emph{Proceedings of the Tenth Workshop on Statistical
  Machine Translation}.\hskip 1em plus 0.5em minus 0.4em\relax Lisbon,
  Portugal: Association for Computational Linguistics, Sep. 2015, pp. 392--395.
  [Online]. Available: \url{https://aclanthology.org/W15-3049}
\BIBentrySTDinterwordspacing

\bibitem{metrics_survey}
\BIBentryALTinterwordspacing
M.~Evtikhiev, E.~Bogomolov, Y.~Sokolov, and T.~Bryksin, ``Out of the bleu: how
  should we assess quality of the code generation models?'' 2022. [Online].
  Available: \url{https://arxiv.org/abs/2208.03133}
\BIBentrySTDinterwordspacing

\bibitem{icarus_verilog}
\BIBentryALTinterwordspacing
S.~Williams, ``Icarus verilog,'' {Version: 10.3}. [Online]. Available:
  \url{http://iverilog.icarus.com/}
\BIBentrySTDinterwordspacing

\bibitem{verilator}
\BIBentryALTinterwordspacing
{Veripool}, ``Verilator,'' {Version: 4.028}. [Online]. Available:
  \url{https://www.veripool.org/verilator/}
\BIBentrySTDinterwordspacing

\bibitem{huggingface}
\BIBentryALTinterwordspacing
T.~Wolf, L.~Debut, V.~Sanh, J.~Chaumond, C.~Delangue, A.~Moi, P.~Cistac,
  T.~Rault, R.~Louf, M.~Funtowicz, J.~Davison, S.~Shleifer, P.~von Platen,
  C.~Ma, Y.~Jernite, J.~Plu, C.~Xu, T.~L. Scao, S.~Gugger, M.~Drame, Q.~Lhoest,
  and A.~M. Rush, ``Transformers: State-of-the-art natural language
  processing,'' in \emph{Proceedings of the 2020 Conference on Empirical
  Methods in Natural Language Processing: System Demonstrations}.\hskip 1em
  plus 0.5em minus 0.4em\relax Online: Association for Computational
  Linguistics, Oct. 2020, pp. 38--45. [Online]. Available:
  \url{https://www.aclweb.org/anthology/2020.emnlp-demos.6}
\BIBentrySTDinterwordspacing

\bibitem{top_k_sampling}
\BIBentryALTinterwordspacing
A.~Fan, M.~Lewis, and Y.~Dauphin, ``Hierarchical neural story generation,''
  2018. [Online]. Available: \url{https://arxiv.org/abs/1805.04833}
\BIBentrySTDinterwordspacing

\end{thebibliography}

\end{document}